\documentclass[twoside,11pt,graphicx,subfigure]{article}

\usepackage{jmlr2e}

\ShortHeadings{Predictive approaches for Gaussian process classifier model selection}{S. Sundararajan and S. Sathiya Keerthi}

\def\bfalp{{\mbox{\boldmath $\alpha$}}}
\def\bfgamma{{\mbox{\boldmath $\gamma$}}}
\def\bfSigma{{\mbox{\boldmath $\Sigma$}}}
\def\bfmu{{\mbox{\boldmath $\mu$}}}

\def\bfbeta{{\mbox{\boldmath $\beta$}}}
\def\bftheta{{\mbox{\boldmath $\theta$}}}

\begin{document}

\title{Predictive Approaches For Gaussian Process Classifier Model Selection \\ September 10, 2008}

\author{\name S. Sundararajan \email ssrajan@yahoo-inc.com \\
       \addr Yahoo! Labs\\
       Bangalore, India\\
       \AND
       \name S. Sathiya Keerthi \email selvarak@yahoo-inc.com \\
       \addr Yahoo! Research \\
       Santa Clara, USA \\
       }

\editor{}

\maketitle

\begin{abstract}
In this paper we consider the problem of Gaussian process classifier (GPC) model selection with different Leave-One-Out (LOO) Cross Validation (CV) based optimization criteria and provide a practical algorithm using LOO predictive distributions with such criteria to select hyperparameters. Apart from the standard average negative logarithm of predictive probability (NLP), we also consider smoothed versions of criteria such as F-measure and Weighted Error Rate (WER), which are useful for handling imbalanced data. Unlike the regression case, LOO predictive distributions for the classifier case are intractable. We 
use 
approximate LOO predictive distributions arrived from Expectation Propagation (EP) approximation. We conduct experiments on several real world benchmark datasets. When the NLP criterion is used for optimizing the hyperparameters, the predictive approaches show better or comparable NLP generalization performance with existing GPC approaches. On the other hand, when the F-measure criterion is used, the F-measure generalization performance improves significantly on several datasets. Overall, the EP-based predictive algorithm comes out as an excellent choice for GP classifier model selection with different optimization criteria.

\end{abstract}

\begin{keywords}
  Gaussian process classification, Model Selection, LOO, Cross Validation, Predictive distributions, Smoothed F-measure, Weighted Error Rate, Precision, Recall, Imbalanced data
\end{keywords}

\section{Introduction}
Gaussian process (GP) models are flexible and powerful probabilistic models that 
are
used to solve classification problems in many areas of application \citep{Raswil:06}. In 
the
Bayesian GP setup, latent function values and hyperparameters that are involved in modeling are integrated with chosen priors. However, the required integrals are often not analytically tractable (due to various choices of likelihoods and priors) and closed form analytic expressions are not available. Two important problems in this context are finding good approximations for integrating over the latent function variables and the hyperparameters. There have been two approaches reported in the literature. In the first approach, both the latent function variables and the hyperparameters are integrated out within some approximations. \citet{Williams:98} used Laplace approximation to integrate over the latent function variables and Hybrid Monte Carlo (HMC) to integrate over the hyperparameters. \citet{Neal:98} used Gibbs sampling to integrate over the latent function variables and HMC to integrate over hyperparameters; this method is more accurate, but it is computationally expensive. In the second approach, only the latent function variables are integrated out and the hyperparameters are optimized on some well-defined objective function. This latter problem of choosing hyperparameters that define the model is essentially the Gaussian process model selection problem and in this paper we focus on this problem.  

There are two commonly used approaches to address this model selection problem. They are marginal likelihood or evidence maximization and minimization of LOO-CV based average negative logarithmic predictive probability (NLP). Both these approaches are available for GP regression model selection 
~\citep{Raswil:06,Sundar:01}. For GP classifier model selection, \citet{Gibbs:00} used a variational approximation method to integrate over the latent function values and estimated the hyperparameters by maximizing marginal likelihood (ML). 
Laplace approximation and Expectation Propagation (EP) approximation ~\citep{Raswil:06,Seeger:03} are other methods that have been used to integrate over the latent function variables. Then, the marginal likelihood is optimized using gradient information obtained with any one of these approximations \citep{Raswil:06,Seeger:03}. \citet{Kim:06} presented an approximate Expectation-Maximization (EM) algorithm to learn the hyperparameters. In the E-step, they used EP to estimate the joint density of latent function values and in the M-step, the hyperparameters were optimized by maximizing 
a variational lower bound on the marginal likelihood. 

In this paper, we consider the approach of using LOO-CV based predictive distributions to address the GP classifier model selection problem. In a related work, \citet{Opper:00} used LOO error estimate to choose rough hyperparameter values by scanning a range of values. In the EP framework \citep{Minka:01}, cavity distributions directly provide LOO error estimates during training and were used to assess predictive performance and select automatic relevance determination (ARD) type hyperparameters \citep{Qi:04}.  The LOO-CV based predictive distributions obtained from probabilistic least squares classifier were used in the minimization of NLP for GP classifier model selection \citep{Raswil:06}. 

In practice, while measures like marginal likelihood and average negative 
logarithmic
predictive probability measures are very useful, other measures like F-measure \citep{van:74} and Weighted Error Rate (WER) are also important and useful, for instance in applications like medical diagnostics, image understanding, etc, where the number of positive examples 
is
much 
smaller
than the number of negative examples. 
Several works that use such measures for hyperparameters optimization exist in the
non-GP literature.
\citet{Hong:07} proposed 
a
kernel classifier construction algorithm based on regularized orthogonal weighted least squares (ROWLS) estimation with LOO-Area Under the ROC Curve (AUC) as model selection criterion for handling imbalanced datasets. \citet{Jansche:05} proposed 
the
training of 
a
probabilistic classifier based on a logistic regression model by optimizing expected F-measure. Here an approximation to F-measure was made so that the F-measure is smoothed and becomes a smooth function of model weights.  
\citet{Seeger:07} proposed a general framework for learning in probabilistic kernel classification models with a large or structured set of classes. He optimized the kernel parameters by minimizing the NLP over k-folds.  \citet{Keerthi:06} considered the task of tuning hyperparameters in SVM models based on minimizing smooth performance validation functions like smoothed k-fold CV error. 
The last three works did not use LOO-CV in their work.


This paper is aimed at addressing two issues. First, the proposed method is different from the work of \citet{Opper:00} and \citet{Qi:04} in that we optimize the hyperparameters directly in the continuous space and any standard non-linear optimization method can be used. It is also different from the LOO-CV based probabilistic LS method \citep{Raswil:06} in that we use the more accurate EP approximation than the LS approximation. Second, criteria such as F-measure and WER, which are needed for tackling imbalanced problems, have not been considered in GP classifier designs.
 
We define smoothed LOO-CV based measures using predictive distributions as a function of GP classifier model hyperparameters. Thus, the objective functions can be optimized using standard non-linear optimization techniques. We investigate usage of LOO-CV predictive distributions obtained from expectation propagation approximation. Actually, the proposed algorithm can also be used with Laplace approximation. However, \citet{Kuss:05} showed for binary classification problems that the EP approximation is better than the Laplace approximation. Therefore, we restrict our attention to using the EP approximation here.  

We conduct experiments on two criteria: the standard average negative logarithm of predictive probability (NLP) and a smoothed version of F-measure.
On the NLP criterion we compare our method (EP-CV(NLP)) against the LOO-CV based probabilistic least squares (LS) classifier \citep{Raswil:06} and standard GP classifier doing ML maximization using EP approximation. We refer the latter two methods as LS-CV(NLP) and EP(ML) respectively; the abbreviations in parentheses refer to the type of objective functions used. The experimental results on several real world benchmark datasets show that the proposed method is better than the LS-CV(NLP) method and is quite competitive to the EP-ML method. On the F-measure criterion we compare the EP-CV(NLP) method with the EP-CV(FM) method. In the latter method we optimize over the F-measure instead of the NLP measure. We also compare with a two-step method\footnote{This method was suggested by an anonymous reviewer.} where, in the first step we optimize over the hyperparameters using the NLP measure and, in the second step we optimize {\it only} the {\it bias} hyperparameter using the F-measure. Experimental results demonstrate that this method is also inferior to the EP-CV(FM) method.   

The 
paper is organized as follows. We give a brief introduction to Gaussian process classification and ML optimization criteria in Section 2. In 
Section 3 we give 
a general set of smooth
LOO-CV based optimization criteria and illustrate their use with LOO-CV predictive distributions. Optimization aspects and 
specific algorithm are given in Section 4. In 
Section 5 we discuss related work on LOO-CV based GPC model selection. In 
Section 6 we present experimental results and 
then conclude 
the paper in Section 7. 
     
\section{Gaussian Process Classification}
In binary classification problems, we are given a training data set ${\it S}$ composed of $n$ input-target pairs $({\bf x}_i, y_i)$ where ${\bf x}_i$ $\in$ $R^D$, $y_i$ $\in$ $\{+1,-1\}$, 
$i \in {\tilde I}$ and ${\tilde I}=\{1,2,\ldots,n\}$. The true function value at ${\bf x}_i$ is represented as a latent variable $f({\bf x}_i)$. The goal is to compute the predictive distribution of the class label $y_*$ at test location ${\bf x}_*$. In standard GPs for classification \citep{Raswil:06}, the latent variables $f({\bf x}_i)$ are modelled as random variables in a zero mean GP indexed by $\{{\bf x}_i\}$. The prior distribution of $\{{\bf f}({\bf X}_n)\}$ is a zero mean multivariate joint Gaussian, denoted as ${\it p}({\bf f})\:=\:{\mathcal N}({\bf 0},{\bf K})$, where ${\bf f}\:=\:[f({\bf x}_1),\ldots,f({\bf x}_n)]^T$, ${\bf X}_n\:=\:[{\bf x}_1,\ldots,{\bf x}_n]$ and ${\bf K}$ is the $n \times n$ covariance matrix whose $(i,j)^{th}$ element is $k({\bf x}_i,{\bf x}_j)$, denoted as ${\bf K}_{i,j}$. One of the most commonly used covariance 
functions
is the squared exponential covariance function given by: $\mathrm{cov}(f({\bf x}_i),f({\bf x}_j))\:=\:k({\bf x}_i,{\bf x}_j)\:=\:\beta_0\:\exp(-{1 \over 2} \sum_{k=1}^D {{(x_{i,k}-x_{j,k})^2} \over {\beta_k}})$. Here, $\beta_0$ represents signal variance and 
the remaining $\beta_k$'s represent width parameters across different input dimensions. Let $\bfbeta\:=\:=[\beta_0, \beta_1, \ldots, \beta_D]$. These parameters are also known as {\it automatic relevance determination} (ARD) hyperparameters. We call this covariance function the ARD Gaussian kernel function. 
Next, it is assumed that the probability over class labels as a function of ${\bf x}$ depends only on the latent function value $f({\bf x})$. For the binary classification problem, given the value of $f({\bf x})$ the probability of class label is independent of all other quantities: $p(y=+1|f({\bf x}),{\it S})\:=\:p(y=+1|f({\bf x}))$ where ${\it S}$ is the dataset. 
The likelihood $p(y_i|f_i)$ can be modelled in several forms 
such as a sigmoidal function or cumulative normal $\Phi(y_i f_i)$ where 
$\Phi(z)\:=\:\int_{-\infty}^z {1 \over {\sqrt{2\pi}}} \exp(-{{w^2} \over 2}) dw$. Assuming that the examples are i.i.d, we have ${\it p}({\bf y}|{\bf f})\:=\:\prod_{i=1}^N p(y_i|f_i;\bfgamma)$. Here, $\bfgamma$ represents hyperparameters that characterize the likelihood.  The prior and likelihood along with the hyperparameters $\bftheta \:=\:[\bfbeta, \bfgamma]$ characterize the GP model. With these modelling assumptions, we can write the 
inference probability 
given $\bftheta$ as:
\begin{equation}
	p(y_*|{\bf x}_*,{\it S},\bftheta)\:=\: \int \:p(y_*|f_*,\bfgamma) \:p(f_*|{\it S},{\bf x}_*,\bftheta)\:\:df_*
\label{pred1}
\end{equation}
Here, the posterior predictive distribution of latent function $f_*$ is given by:
	$$p(f_*|{\it S},{\bf x}_*,\bftheta)\:=\:\int\:p(f_*|{\bf x}_*,{\bf f},\bfbeta) \:p({\bf f}|{\it S},\bftheta)\:d{\bf f}.$$
where $p({\bf f}|{\it S},\bftheta) \varpropto \prod_{i=1}^N p(y_i|f_i,\bfgamma)\: p({\bf f}|{\bf X},\bfbeta)$. In a Bayesian solution, the class probability at the test point $x_*$ would be obtained by integrating over the hyperparameters weighted by their posterior probability 
$$ p(y_*|x_*,{\it S})\:=\:\int \:p(y_*|x_*,{\it S},\bftheta)\:p(\bftheta|{\it S}) \:d\bftheta.$$
In general there is no closed form expression available for this integral and 
it
is expensive to compute. Therefore, instead of integrating over the hyperparameters, 
a single set of their values
is  
usually
estimated from the dataset by optimizing various criteria as mentioned earlier and 
then
used in (\ref{pred1}). 

\subsection{Marginal Likelihood Maximization}
Marginal likelihood or evidence maximization \citep{Raswil:06} is commonly used to estimate the hyperparameters during model selection. The marginal likelihood is given by: 
\begin{equation}
	p({\bf y}|{\bf X},\bftheta)\:=\:\int\:\prod_{i=1}^N p(y_i|f_i,\bfgamma)\: p({\bf f}|{\bf X},\bfbeta)\:d{\bf f}
\label{ML}
\end{equation}
This integral cannot be calculated analytically except for a special case like GP regression with Gaussian noise. Therefore, certain approximations are needed to compute these quantities. Laplace approximation and EP approximations are two popular methods used for this purpose. To gain more insight into the quality of the Laplace and EP approximations, \citet{Kuss:05} carried out comprehensive comparisons of these approximations with (the more exact) Markov Chain Monte Carlo (MCMC) sampling approach in terms of their predictive performance and marginal likelihood estimates. They found that EP is the method to be used for approximate inference in binary GPC models, when the computational cost of MCMC is prohibitive. Hence in our study we restrict ourselves to the more accurate EP approximation given in the next subsection. With the EP approximation the hyperparameters are learnt by optimizing marginal likelihood with gradient information \citep{Raswil:06, Seeger:03} using standard non-linear optimization techniques. 

\subsection{Expectation Propagation}
The EP algorithm is an iterative algorithm which is used for approximate Bayesian inference \citep{Opper:00,Minka:01}. It has been applied to GP classification \citep{Raswil:06, Seeger:03} 
. EP finds a Gaussian approximation $q({\bf f}|{\it S},\bftheta)\:=\:\mathcal{N}({\bf f}|{\bf m},{\bf C})$ to the posterior $p({\bf f}|{\it S},\bftheta)$ by moment matching of approximate marginal distribution and the posterior. Mean and covariance of the approximate Gaussian are given by: 
\begin{equation}
{\bf m}\:=\:{\bf C}\bfSigma^{-1}{\bfmu} \:\:\:\:\:\: {\bf C}\:=\:({\bf K}^{-1}\:+\:\bfSigma^{-1})^{-1}
\label{posmean}
\end{equation}
where 
${\bfmu}\:=\:(\mu_1,\mu_2,\ldots,\mu_n)^T$ and $\bfSigma\:=\:diag(\sigma^2_1,\sigma^2_2,\ldots,\sigma^2_n)$ are called site function parameters. As per the approximation, the posterior is written in terms of the site functions $t(f_i;\mu_i,\sigma^2_i,Z_i)\:=\:Z_i {\mathcal N}(f_i|\mu_i,\sigma^2_i)$ and prior $p({\bf f}|{\bf X},\bftheta)$ as 
$$ q({\bf f}|{\it S},\bftheta)\:=\:{{p({\bf f}|{\bf X},\bftheta)} \over q({\it S}|\bftheta)} \prod_{i=1}^N\:t(f_i;\mu_i,\sigma^2_i,Z_i).$$ The EP algorithm iteratively visits each site function in turn, and adjusts the site parameters to match moments of an approximation to the posterior marginals. This process requires replacement of intractable exact {\it cavity} distribution with a tractable approximation based on the site functions and is given by:
$$q_{\setminus{i}}(f_i)\:=\:\int \prod_{j\neq i} t(f_j;\mu_j,\sigma^2_j,Z_j) \:p({\bf f}|{\bf X},\bftheta)\:d{\bf f}^{\setminus{i}}$$ where $q_{\setminus{i}}(f_i) \varpropto \mathcal{N}(f_i|\mu_{\setminus{i}},\sigma^2_{\setminus{i}})$ is the approximate cavity function and is related to 
the
diagonal entries of the posterior $q({\bf f}|{\it S},\bftheta)$ as:
$ q_{\setminus{i}}(f_i)t(f_i;\mu_i,\sigma^2_i,Z_i) \:\varpropto \:\mathcal{N}(f_i|m_i,{\bf C}_{ii}).$ Here, 
the
${\bf C}_{ii}$ 
represent 
the diagonal entries of the matrix ${\bf C}$. Using Gaussian identities, the mean and variance of cavity distribution is related to the site parameters as:
\begin{equation}
	\mu_{\setminus{i}}\:=\:\sigma^2_{\setminus{i}}\bigl({m_i \over {\bf C}_{ii}}-{\mu_i \over \sigma^2_i}\bigr) \:\:\:\:\:\:\:\:\:\sigma^2_{\setminus{i}}\:=\:\bigl(({\bf C}_{ii})^{-1}-\sigma^{-2}_{i}\bigr)^{-1}
\label{cavmu}
\end{equation}              
Then, EP adjusts the site parameters $\mu_i$, $\sigma^2_i$ and $Z_i$ such that the approximate posterior marginal using the exact likelihood approximates the posterior marginal based on the site function well. That is, $q_{\setminus{i}}(f_i)\:p(y_i|f_i) \simeq q_{\setminus{i}}(f_i)\:t(f_i;\mu_i,\sigma^2_i,Z_i)$. This is done by matching the zeroth, first and second moments on both sides. Thus, the EP algorithm iteratively updates site parameters until convergence.  
Though there is no convergence proof, 
in practice
the EP algorithm converges 
in most cases. See \citet{Raswil:06} for more details. 

Next, within some constant, the marginal likelihood with EP approximation \citep{Raswil:06} is given by: 
\begin{equation}
	\log q({\bf y}|{\bf X},\bftheta)\:=\:-{1\over2}\log|{\bf K}+\bfSigma|\:-{1\over 2}\bfmu^T({\bf K}+\bfSigma)^{-1}\bfmu\:+\:\sum_{i=1}^n\log w_i
\label{EPML}
\end{equation}
where $w_i=\Phi(z_i)\exp\Bigl({{(\mu_{\setminus{i}}-\mu_i)^2} \over {2(\sigma^2_{\setminus{i}}+\sigma^2_i)}}\Bigr)\sqrt{\sigma^2_{\setminus{i}}+\sigma^2_i}$ and $z_i={{y_i \mu_{\setminus{i}}} \over \sqrt{{1+\sigma^2_{\setminus{i}}}}}$. 
The 
hyperparameters are optimized using gradient expressions with standard conjugate gradient or quasi-Newton type non-linear optimization techniques.  

\section{Leave-One-Out Cross Validation based Optimization Criteria}
In this section, we 
give definitions of various LOO-CV based optimization criteria. 
In section 4
we give details on how these measures can be optimized using standard nonlinear optimization 
techniques. 
The LOO predictive distributions  $p(y_i|{\bf x}_i,{\it S}_{\setminus{i}},\bftheta)$, $i\in\tilde I$ 
play a crucial role. 
Here
${\it S}_{\setminus{i}}$ represents the dataset without $i$th example.
Their exact computation is expensive. In section 4 we will also discuss how to approximate them efficiently.


\subsection{NLP Measure}
The averaged negative logarithm of predictive probability (NLP) is defined as: 
\begin{equation}
	G(\bftheta)\:=\:-{1\over n}\sum_{i=1}^n \:\log\:p(y_i|x_i,{\it S}_{\setminus{i}},\bftheta)
\label{nlp}
\end{equation}
This 
LOO-CV based
measure is generic and has been used in the context of probabilistic LS classifiers 
(see section 5)
and GP regression \citep{Raswil:06, Sundar:01}. 

     
\subsection{Smoothed LOO Measures}
While 
measures such as
marginal likelihood (\ref{EPML}) 
and NLP in (\ref{nlp}) are useful for normal situations, 
other measures like F-measure 
and WER  
are important, for example, 
when dealing with
imbalanced datasets. 
Let us now define these measures.

Consider the binary classification problem with class labels +1 and -1. Assume that there are $n_+$ positive examples and $n_{-}$ negative examples. In general, the performance of the classifier may be evaluated using counts of data samples $\{a, b, c, d\}$ defined via the confusion matrix given in Table 1. 
Let $n_{+}\:=\:a+b$ and $n_{-}\:=\:c+d$. The true positive rate (TP) is the proportion of positive data samples that were correctly classified (that is, true positives) and the false positive rate (FP) is the proportion of negative data samples that were incorrectly classified (that is, false positives) \citep{Hong:07}. These rates are given by: TP$\:=\:{a \over {a+b}}\:=\:{a \over n_{+}}$ and FP$\:=\:{c \over{c+d}}\:=\:{c \over {n_{-}}}$. 
The misclassification rate is given by: MCR$\:=\:{{b+c} \over n}$. Note that the true positive rate is also known as {\it Recall} (R). {\it Precision} is another important quantity defined as: P$\:=\:{a \over {a+c}}$. 

Now let us consider the imbalanced data case and assume that $n_{-}\gg n_{+}$. In this case if MCR is minimized then the classifier will be biased toward the negative class due to its effort in minimizing the false positives (that is c) more strongly than minimizing false negatives (that is b). In the worst case almost all the positive examples will be wrongly classified, that is $a\rightarrow0$ . This results in both $P\rightarrow 0$ and $R\rightarrow 0$. Thus MCR is not a good measure to use when the dataset is imbalanced. This problem can be addressed by optimizing other measures that we discuss next. 

The F-measure is one such measure and is defined \citep{van:74} as: 
	$$F_{\zeta}(P,R)\:=\:\Bigl({\zeta \over R}+{{1-\zeta}\over P}\Bigr)^{-1}$$
\begin{table}
\caption{Confusion Matrix for Binary classification}
\vskip 0.05in
\begin{small}
\begin{center}
\begin{tabular}{|l|c|c|} \hline
 & Positive (Predicted) & Negative (Predicted)  \\ \hline
Positive (Actual) & a & b \\ \hline 
Negative (Actual)  & c & d \\ \hline 
\end{tabular}
\end{center}
\end{small}
\end{table}
where $0\le \zeta \le 1$ and $0\le F_{\zeta}(P,R)\le 1$. It has been used in various applications like document retrieval \citep{van:74} and text classification \citep{Joachims:05}. It is particularly preferable over MCR when the dataset is highly imbalanced \citep{Joachims:05,Jansche:05}. 

Let us get into more details on the functioning of the F-measure. When $\zeta \rightarrow 0$, we get $F_{\zeta}(P,R) \rightarrow P$. Then optimizing the F-measure means we are interested {\it only} in maximizing the {\it Precision}. On the other hand, when $\zeta \rightarrow 1$, we get $F_{\zeta}(P,R) \rightarrow R$. In this case we are interested {\it only} in maximizing the {\it Recall}. The user can choose an appropriate value for $\zeta$ depending on how much of importance he/she wants to give to the precision and recall. Thus, the F-measure combines precision and recall into a single optimization criterion by taking their $\zeta$-weighted harmonic mean. In the imbalanced data case mentioned above MCR minimization can potentially result in $F_{\zeta}(P,R)\rightarrow 0$. By maximizing the F-measure we can prevent the classifier from being completely biased towards the negative class. Note that $F_{\zeta}(P,R)$ can be re-written in terms of $a$, $b$ and $c$ as:
\begin{equation}
	F_{\zeta}(a,b,c)\:=\:{a \over {a+\zeta b+(1-\zeta) c}}
\label{Fabcmeasure}
\end{equation} 
In all our experiments we set $\zeta\:=\:0.5$. In this case, it becomes $F_{0.5}(P,R)\:=\:{{2PR} \over {P+R}}$ and can also be written as: $F_{0.5}(a,b,c)\:=\:{1 \over {1+{{b+c}\over {2a}}}}$.  Then, maximizing $F_{0.5}(a,b,c)$ is equivalent to maximizing ${a} \over {b+c}$. Thus, we can maximize $F_{0.5}(a,b,c)$ by both minimizing the error (that is, $b+c$) and maximizing the true positives. The trade-off kicks-in since maximizing the true positives tends to increase the false positives. 
Thus maximizing the F-measure controls both the true positives and the error appropriately. In general the F-measure summarizes a classifier's ability to identify the positive class and plays an important role in the evaluation of binary classifier. As a criterion for optimizing hyperparameters 
F-measure can be computed 
on
an evaluation or validation dataset. 
However, in practical situations involving small datasets\footnote{GP models are known to be particularly valuable for problems with small datasets.} it is wasteful to employ a separate evaluation set.
The
LOO-CV approach would be useful in such situations and we show next how the F-measure can be estimated with such approach. 

\citet{Hong:07} estimated $TP$ and $FP$ as:
	$$\widehat{TP}\:=\:{1 \over {n_+}} \sum_{i=1}^n T({\hat y}_{\setminus{i}}y_i,y_i),$$
	$$\widehat{FP}\:=\:{1 \over {n_-}} \sum_{i=1}^n F({\hat y}_{\setminus{i}}y_i,y_i).$$
Here ${\hat y}_{\setminus{i}}$ represents predicted label for $i$th sample. Therefore ${\hat y}_{\setminus{i}}y_i$ takes value $+1$ when the prediction matches with the actual label and $-1$ otherwise. $T(u,v)$ is an indicator function which is 1 if $u=1$ and $v=1$. Similarly, $F(u,v)$ is one if $u=-1$ and $v=-1$. Otherwise, these functions take zero values. 
\citet{Hong:07} 
used these estimates to compute $\widehat{AUC}\:=\:{{1+\widehat{TP}-\widehat{FP}} \over 2}$ as an approximation of AUC and used this criterion to select 
a
subset of basis vectors in 
a
kernel classifier model construction procedure for imbalanced datasets. Note that this definition of AUC is applicable only for a hard classifier (fixed non-probabilistic classifier) with binary outputs. See \citet{Hong:07} for more details. In a strict sense such a definition of AUC is not suitable for a probabilistic classifier like GP classifier that provides continuous probabilistic output. 
However, we can make use of this approach of defining $TP$ and $FP$ as above to compute the quantities $a$, $b$ and $c$ that are needed to evaluate the F-measure 
in (\ref{Fabcmeasure}). There are two issues associated with these estimates. The first issue is 
that
these estimates are not 
smooth (in fact, not even continuous) 
functions
of hyperparameters. Therefore they cannot be used directly in any  
approach that uses gradient-based nonlinear optimization methods to tune the hyperparameters.
Secondly, these estimates do not use predictive probability values which is particularly important when we want to take variance also into account. 
In non-GP contexts
\citet{Jansche:05} and \citet{Keerthi:06} addressed the first issue by defining smoothed F-measure or 
other
validation functions by replacing the indicator function with a sigmoid  
function,
which makes the optimization criterion as a 
smooth
function of hyperparameters. However, they did not consider 
a
LOO approach
and used a validation set instead.
\citet{Jansche:05} considered maximum {\it a posterior} probabilities and \citet{Keerthi:06} used sigmoidal approximations for SVM models. Here, we propose to combine LOO based estimation and 
smoothed version of the quantities $\{a, b, c, d\}$ 
denoted as $A(\bftheta)$, $B(\bftheta)$, $C(\bftheta)$ and $D(\bftheta)$.
We can set
\begin{equation}
	A(\bftheta)\:=\: \sum_{i:y_i=+1}\:p(y_i=+1|x_i,{\it S}_{\setminus{i}},\bftheta)
\label{Atheta}
\end{equation}
Since $n_{+}=a+b$, we can write $B(\bftheta)\:=\:n_{+}-A(\bftheta)$. With $m_{+}$ denoting the number of examples predicted as positive, we can parameterize it as $m_{+}(\bftheta)\:=\:A(\bftheta)+C(\bftheta)$. This can be rewritten as: 
\begin{equation}
m_{+}(\bftheta)\:=\:\sum_{i=1}^n p(y_i=+1|x_i,{\it S}_{\setminus{i}},\bftheta)
\label{Mtheta}
\end{equation}
Thus, the smoothed F-measure can be defined from (\ref{Fabcmeasure}) as: 
\begin{equation}
	F_{\zeta}(\bftheta)\:=\:{{A(\bftheta)} \over {\zeta n_{+} + (1-\zeta) m_{+}(\bftheta)}}
\label{FMmooth}
\end{equation} 
Note that $D(\bftheta)$ can be defined in a similar fashion as $m_{-}(\bftheta)\:=\:B(\bftheta)+D(\bftheta)$. Using these quantities, other derived quantities like $TP(\bftheta)$ and $FP(\bftheta)$ can be defined as LOO based estimates. Then, smoothed LOO estimates of WER can be obtained as shown below.

The WER measure is another useful measure for imbalanced datasets. Using the quantities defined above, its smoothed version can be written as:
\begin{equation} 
WER(\bftheta;\tau)\:=\:{{n_{+}(1-TP(\bftheta))+\tau n_{-} FP(\bftheta)} \over {n_{+}+\tau n_{-}}}.
\label{WER}
\end{equation}
where $\tau$ is the ratio of the cost of mis-classifications of the negative class to that of the positive class and $0\le \tau \le 1$. Thus by choosing a suitable $\tau$ value for a given problem and optimizing over the hyperparameters we can design classifiers without becoming biased toward one class. Note that for ease of notation we have omitted hat on $TP(\cdot)$ and $FP(\cdot)$. 
Following the work of \citet{Hong:07}  one can also define 
\begin{equation}
AUC(\bftheta)\:=\:{{1+TP(\bftheta)-FP(\bftheta)} \over 2}
\label{AUC}
\end{equation}
 and optimize over the hyperparameters. As mentioned earlier, such a definition is not suitable for the GP classifier. Nevertheless it is interesting to note that it has the desirable property of trading-off between high TP and low FP. Also, on comparing this definition of AUC with (\ref{WER}) we see that they are related in the sense that maximizing AUC is same as minimizing WER when $\tau=1$ {\it and} $n_{+}\:=\:n_{-}$. 

Overall we see that the LOO-CV predictive distributions can be used to define various criteria that are 
smooth
functions of hyperparameters resulting in smoothed LOO-CV measures. Now given that the LOO-CV predictive distributions are readily available from the EP algorithm, we can optimize the various smoothed LOO-CV measures directly using standard non-linear optimization techniques.  

\section{EP-CV Algorithm for Choosing Hyperparameters}
Various criteria 
such as 
(\ref{nlp}), (\ref{FMmooth}), (\ref{WER}) and (\ref{AUC})  
depend 
on the hyperparameters $\bftheta$ via the predictive distributions 
$p(y_i|x_i,{\it S}_{\setminus{i}},\bftheta)$.
With cumulative Gaussian likelihood, they can be written as:
\begin{equation}
	p(y_i|x_i,{\it S}_{\setminus{i}},\bftheta)\:=\:\Phi\Bigl({{y_i(\mu_{\setminus{i}}+\gamma)} \over {\sqrt{1+\sigma^2_{\setminus{i}}}}}\Bigr)
\label{Pyi}
\end{equation}
Note that (\ref{Pyi}) is obtained from (\ref{pred1}) with $p(f_{i}|x_i,S_{\setminus{i}},\bftheta)=\mathcal{N}(\mu_{\setminus{i}},\sigma^2_{\setminus{i}})$. The hyperparameter $\gamma$ is referred to as the bias parameter and it helps in shifting the decision  
boundary with the probability value ${1\over2}$. In general, the bias hyperparameter $\gamma$ is very useful \citep{Raswil:06, Seeger:03} and can be optimized. 

\subsection*{EP Approximation}
To compute (\ref{Pyi}) we need the LOO mean $\mu_{\setminus{i}}$ and variance 
$\sigma^2_{\setminus{i}}$ $\forall i$. 
With the EP approximation, they can be computed using (\ref{cavmu}). 
Full details of gradient calculations needed for implementing hyperparameter optimization are given in the appendix.
We take the expectation-maximization (EM) type approach for hyperparameters optimization.  
 This is because gradient expressions involving implicit derivatives (with site parameters varying as a function of hyperparameters) are not available due to the iterative nature of 
the EP algorithm. This approach results in the following algorithm.             
\subsubsection*{\bf {EP-CV Algorithm}:}
\begin{enumerate}
\item Initialize the hyperparameters $\bftheta.$
\item {\bf Perform E-Step}: Given the hyperparameters, we find the site parameters  ${\bfmu}$ and $\Sigma$ and the posterior $q({\bf f}|{\it S},\bftheta)\:=\:\mathcal{N}({\bf m},{\bf C})$ using EP algorithm.
\item {\bf Perform M-Step}: Find the hyperparameters $\bftheta$ by optimizing over any LOO-CV based measure like (\ref{nlp}), (\ref{FMmooth}), (\ref{WER}) or (\ref{AUC}) using any standard gradient based optimization technique. We carry out just one line search in this optimization process. During this line search as the hyperparameters change, we perform the following sequence of operations. 
\begin{enumerate}
\item Compute the posterior mean ${\bf m}$ and covariance ${\bf C}$ using (\ref{posmean}).
\item Compute the LOO mean $\mu_{\setminus{i}}$ and variance $\sigma^2_{\setminus{i}}$ using (\ref{cavmu}).
\item Compute the chosen objective function like (\ref{nlp}), (\ref{FMmooth}), (\ref{WER}) or (\ref{AUC})  and its derivatives.
\end{enumerate}
Note that through out this M-step, it is assumed that the site parameters are fixed and the values obtained from step (2) are used.   
\item repeat steps 2-3 until there is no significant change in the objective function value.
\end{enumerate}   
This algorithm 
worked
well in our experiments. 
A similar 
EM approach was used by \citet{Kim:06} (which they called EM-EP algorithm) in the optimization of a lower bound on the marginal likelihood. 

\begin{figure}
{
	\includegraphics[width=7cm]{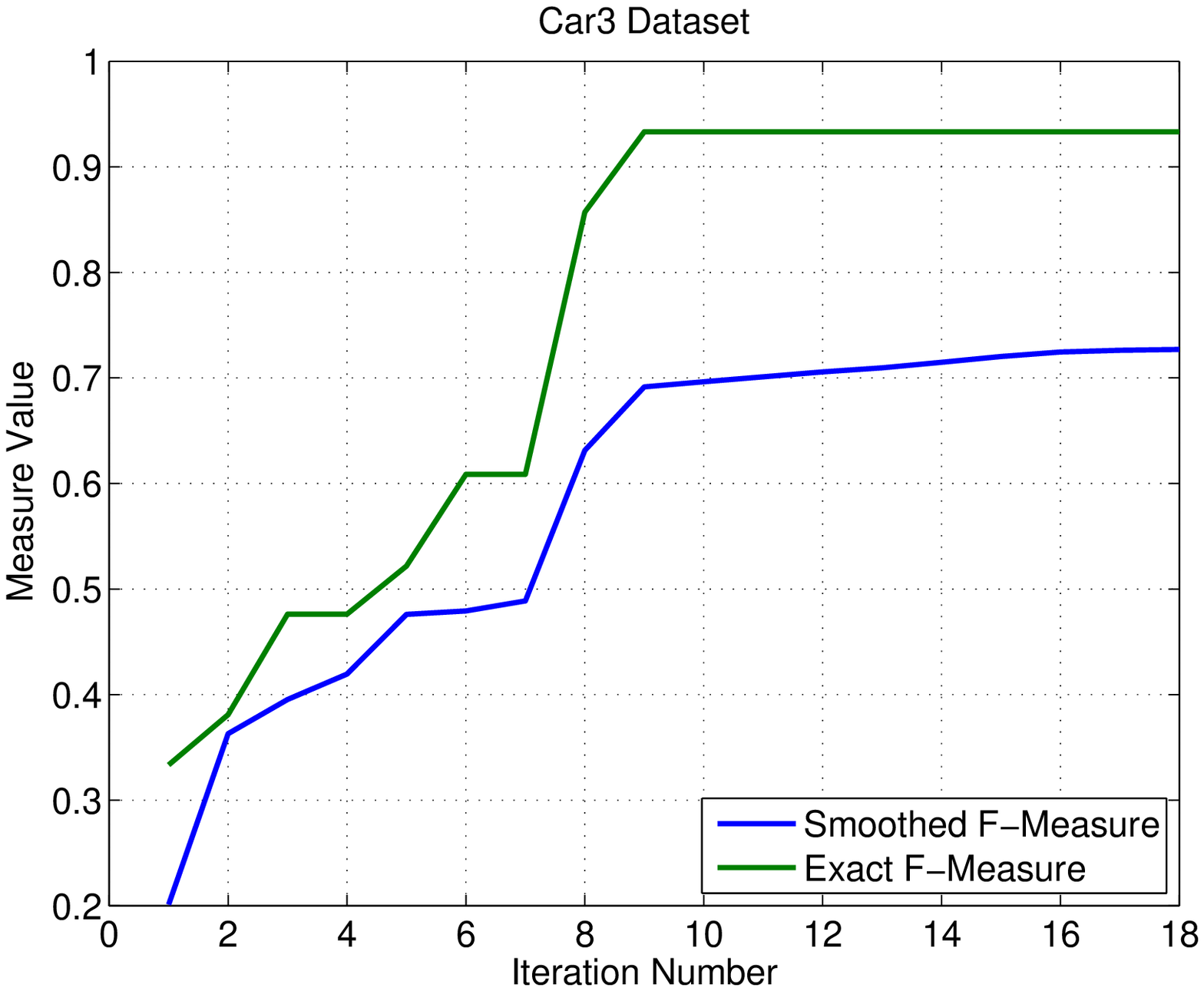}
}
\hspace{1cm}
{
	\includegraphics[width=7cm]{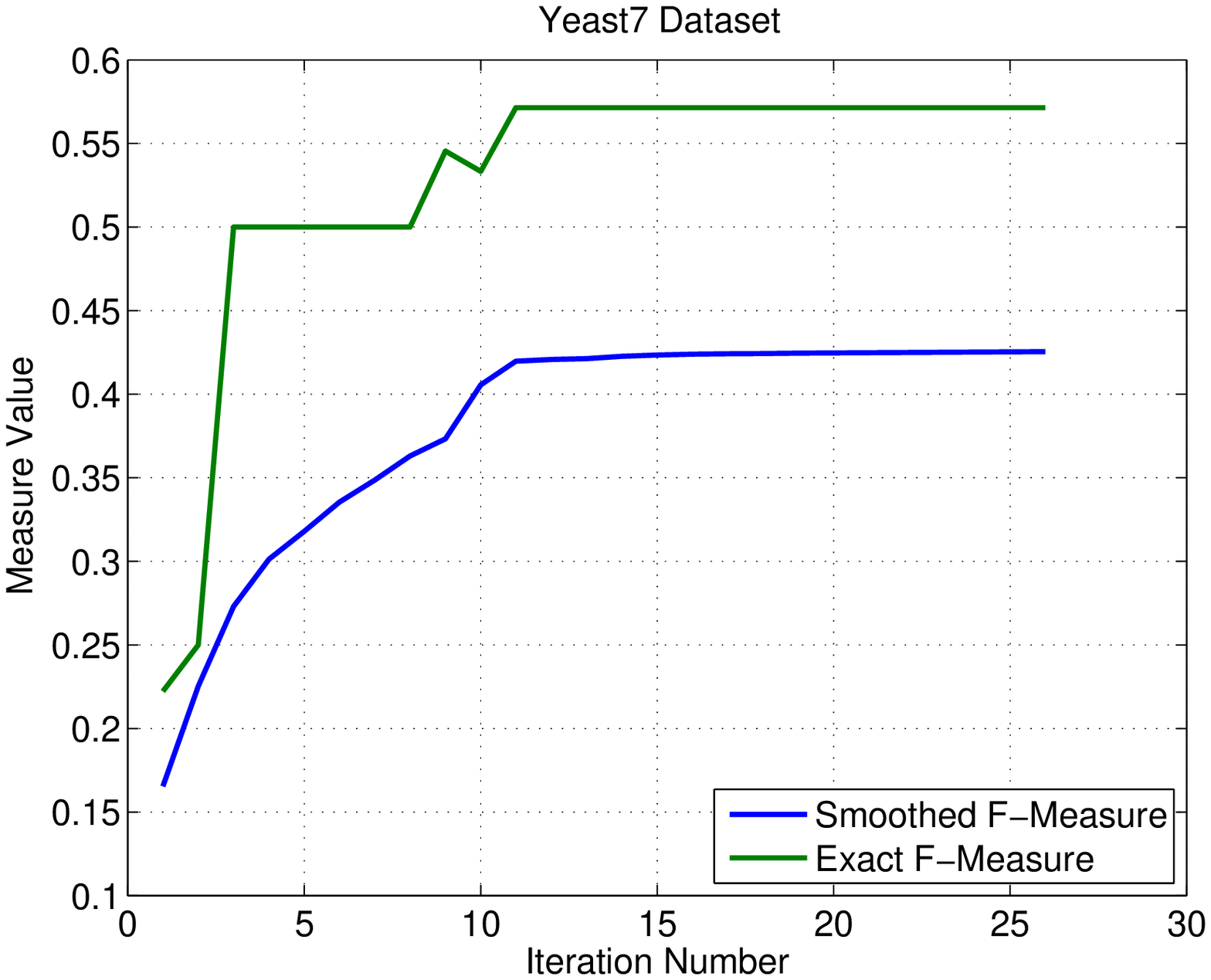}

}
\caption{F-Measure Optimization on Car3 and Yeast7 Datasets}
\label{fig1}
\end{figure}
Since the EP-CV algorithm optimizes the smoothed F-measure it is useful to study the behavior of the true F-measure as optimization proceeds. We do this study on two of the datasets described in Table 6 of section 6. The optimization algorithm was terminated when there was no significant change in the smoothed F-measure value. From Figure 1 we see that the smoothed F-measure monotonically increases in both cases. Also as expected, the true F-measure exhibits a non-smooth behavior expected of a discrete function, and also, the values of true and smoothed F-measures are not the same. The difference arises because the smoothed F-measure is based on probabilistic scores, which can take any value between 0.5 and 1 depending on the problem (even when correct classification occurs). The important point to observe is that, in general, there is an increasing trend in true F-measure value as the optimization progresses. In the case of Car3 dataset (left panel) we see that clearly. A similar trend is seen in the case of Yeast7 dataset (right panel) also, except for a small dip at the 10th iteration. Though such a behavior happens sometimes in early iterations, we observed that better true F-measure value is almost always obtained as the optimization progresses.         
\subsection*{Computational and Storage Complexities}
The computational complexity of the EP-CV algorithm depends on the 
number of ARD kernel parameters $D$. See appendix for more details. For a given problem with $D$ fixed, the complexity is $O(n^3)$. This complexity is same as that of the EP(ML) method (see equation (\ref{EPML})) and LS-CV(NLP) method given in the next section. 
Also in many practical problems a single global scaling hyperparameter for all the input features is sufficient. Finally, the storage complexities of all the methods are $O(n^2)$.       

\section{Other LOO-CV based Methods}
Having discussed our approach in detail it is useful to recall and discuss other LOO-CV based GPC model selection methods in relation to it.

\citet{Opper:00} derived a mean field algorithm for binary classification with GPs based on the TAP approach originally proposed in 
the
statistical physics of disordered systems. They showed that this approach yields an approximate LOO estimator for the generalization error. 
This estimate is equivalent to the 
LOO-CV
error estimate obtained from EP \citep{Minka:01}. 
Instead 
of optimizing over the hyperparameters, 
\citet{Opper:00} used the LOO-CV error count (using indicator functions) to choose rough hyperparameter values by scanning a range of values. 

\citet{Qi:04} used the LOO-CV error estimate obtained from EP to determine ARD hyperparameters. They worked with the Gaussian process classifier where each input feature is associated with a weight parameter $v_i$ with the prior $\mathcal{N}(0,{\zeta_i}^{-1})$. The hyperparameters $\zeta_i^{-1}$
were obtained by 
{\it maximizing the evidence}
using a fast sequential update based on the work of \citet{Faul:02}. The outcome of this optimization is that many $\zeta_i$s would go to infinity such that only a few nonzero weights $v_i$ will be present. 
Even though the ARD hyperparameters were optimized by maximizing the evidence, 
to prevent
overfitting 
\citet{Qi:04} proposed to select the final model as the one that gives the minimum LOO-CV error count or probability. As the LOO-CV error count is discrete, they chose the model with maximum evidence when there is a tie in the count. Compared to this approach, we work with 
the
GP classifier model (without the weight parameters) detailed in Section 2 and optimize over the hyperparameters (including ARD) directly with various LOO-CV based measures (including F-measure, WER etc.) using gradient information. 

In this context, 
the
LOO-CV based probabilistic LS classifier \citep{Raswil:06} is a 
more
direct LOO-CV based GPC model selection approach. 
For the sake of completion we give some details here and 
later
compare our algorithms with this approach 
in our experiments. This approach treats classification as a regression problem. 
Note that the probabilistic interpretation of LS criterion implies 
a
Gaussian noise model. But the output ${\bf y}$ can take only $+1$ or $-1$ which is 
slightly
odd. However, this approach is simple to implement and a probabilistic interpretation is given by passing the predictions through a sigmoid. 

Specifically, the LOO mean $\mu_{\setminus{i}}$ and variance $\sigma^2_{\setminus{i}},\:i\in {\tilde I}$ are obtained from LOO-CV formulation of GP regression and the predictive distributions are obtained 
via
(\ref{Pyi}). Here, $\mu_{\setminus{i}}$ and $\sigma^2_{\setminus{i}}$ are given by:
\begin{equation}
	\mu_{\setminus{i}}\:=\:y_i\:-\:{\tilde \alpha}_i \sigma^2_{\setminus{i}}  \:\:\:\:\:\:\sigma^2_{\setminus{i}}\:=\:{1 \over  {{\bar {\bf K}}_{ii}}}
\label{lsmu}
\end{equation}
where ${\tilde \bfalp}\:=\:{\bar {\bf K}}{\bf y}$ and ${\bar {\bf K}}\:=\:({\bf K}+\lambda{\bf I})^{-1}$. 
Here $\lambda$ can 
either
be set to a small 
positive
value or 
treated as a regularization hyperparameter with 
a small upper bound 
constraint. This is useful when ${\bf K}$ can become ill-conditioned during optimization. Finally, the hyperparameters are optimized using 
(\ref{nlp}). We call this method as LS-CV(NLP).


\begin{table}
\caption{Various methods and their descriptions. All the EP-CV methods are optimized using EP-CV algorithm.}
\vskip 0.05in
\begin{small}
\begin{sc}
\begin{center}
\begin{tabular}{|l|l|} \hline
Method & Description  \\ \hline
EP(ML)  & Marginal likelihood maximization within EP approximation. \\
        & That is, optimize (\ref{EPML}) over $\bftheta$. \\ \hline 
EP-CV(NLP)  & Negative Logarithmic Predictive loss minimization within EP approximation. \\
            & That is, optimize (\ref{nlp}) over $\bftheta$. For ease of notation, this \\
            & method is referred as {\bf NLP} in Table 7. \\ \hline 
LS-CV(NLP)  & Negative Logarithmic Predictive loss minimization within Least Squares \\
            & approximation as described in Section 5. Optimize (\ref{nlp}) over $\bftheta$. \\ \hline 
EP-CV(FM)  & F-Measure maximization within EP-CV approximation. \\ 
           & That is, optimize (\ref{FMmooth}) over $\bftheta$. For ease of notation, this \\
           & method is referred as {\bf FM} in Table 7. \\ \hline 
NLP-FM(BIAS)     & In the first step, hyperparameters are optimized using EP-CV(NLP) method. \\
           & In the second step, only the bias parameter ($\gamma$) is optimized using \\ 
	   & EP-CV(FM) method.We also refer this method as two-step method. \\ \hline 
\end{tabular}
\end{center}
\vskip -0.2in
\end{sc}
\end{small}
\end{table}

\begin{table}
\caption{Data sets description: NLP Experiment. Here, $n$, $D$, $p$ and $nr$ represent the numbers of training examples, input dimension, test examples and train/test partitions respectively.}
\vskip 0.05in
\begin{small}
\begin{sc}
\begin{center}
\begin{tabular}{|l|l|l|l|l|} \hline
Dataset & $n$ & $D$  & $p$ & $nr$ \\ \hline
Banana  & 400 & 2 & 4900 & 100 \\ \hline 
Breastcancer  & 200 & 9 & 77 & 100 \\ \hline 
Diabetes  & 468 & 8 & 300 & 100  \\ \hline 
German & 700 & 20 & 300 & 100  \\ \hline
Heart  & 170 & 13 & 100 & 100  \\ \hline 
Image & 1300 & 18 & 1010 & 20  \\ \hline
Ringnorm & 400 & 20 & 7000 & 100  \\ \hline
Splice & 1000 & 60 & 2175 & 20 \\ \hline
Thyroid & 140 & 5 & 75 & 100 \\ \hline
Titanic & 150 & 3 & 2051 & 100 \\ \hline
Twonorm & 400 & 20 & 7000 & 100 \\ \hline
Waveform & 400 & 21 & 4600 & 100 \\ \hline
\end{tabular}
\end{center}
\vskip -0.2in
\end{sc}
\end{small}
\end{table}


\begin{table}
\begin{center}
\caption{NLP Performance} 
\vskip 0.1in
\begin{small}
\begin{sc}
\begin{tabular}{|l|c|c|c|} \hline
 Dataset/Method & EP(ML) & EP-CV(NLP)& LS-CV(NLP)\\ \hline
Banana & 23.90 $\pm$ 0.81 & 24.26 $\pm$ 1.06 & 33.88 $\pm$ 1.89 \\ \hline
Breastcancer & 53.57 $\pm$ 4.75 & 54.12 $\pm$ 5.27 & 55.58 $\pm$ 5.03 \\ \hline 
Diabetes & 47.74 $\pm$ 1.96 & 47.97 $\pm$ 2.09 & 50.72 $\pm$ 2.13 \\ \hline
German & 48.67 $\pm$ 2.74 & 49.05 $\pm$ 2.76 & 50.51 $\pm$ 2.30 \\ \hline 
Heart &  40.16 $\pm$ 5.36 & 40.03 $\pm$ 5.00 & 45.11 $\pm$ 4.91 \\ \hline
Image & 8.26 $\pm$ 1.07 & 8.45 $\pm$ 0.97 & 22.70 $\pm$ 0.66 \\ \hline
Ringnorm & 16.88 $\pm$0.93 & 16.56 $\pm$ 1.01 & 28.48 $\pm$ 0.75 \\ \hline 
Solar & 57.25 $\pm$ 1.38 & 57.35 $\pm$ 1.42 & 59.61 $\pm$ 1.34 \\ \hline
Splice & 28.48 $\pm$ 0.88 & 29.60 $\pm$ 0.79 & 36.83 $\pm$ 0.42 \\ \hline
Thyroid & 10.21 $\pm$ 3.76 & 9.94 $\pm$ 3.69 & 25.33 $\pm$ 4.86 \\ \hline
Titanic & 66.86 $\pm$ 1.97 & 51.73 $\pm$ 1.73 & 53.78 $\pm$ 14.08 \\ \hline
Twonorm & 8.31 $\pm$ 0.88  & 9.08 $\pm$ 1.97 & 25.94 $\pm$ 0.53 \\ \hline
Waveform & 23.01 $\pm$ 0.89 & 22.97 $\pm$ 0.67 & 32.63 $\pm$ 0.59 \\ \hline
\end{tabular}
\end{sc}
\end{small}
\end{center}
\end{table}

\begin{table}
\begin{center}
\caption{Test Set Error Performance} 
\vskip 0.1in
\begin{small}
\begin{sc}
\begin{tabular}{|l|c|c|c|} \hline
 Dataset/Method & EP(ML) & EP-CV(NLP)& LS-CV(NLP)\\ \hline
Banana & 10.41 $\pm$ 0.65 & 10.51 $\pm$ 0.50 & 10.93 $\pm$ 0.67 \\ \hline
Breastcancer & 26.52 $\pm$ 4.89 & 26.61 $\pm$ 4.80 & 25.94 $\pm$ 4.59 \\ \hline 
Diabetes & 23.28 $\pm$ 1.82 & 23.41 $\pm$ 1.82 & 24.30 $\pm$ 2.51 \\ \hline
German & 23.36 $\pm$ 2.11 & 23.48 $\pm$ 2.00 & 23.94 $\pm$ 2.33 \\ \hline 
Heart & 16.65 $\pm$ 2.87 & 16.62 $\pm$ 3.08 & 17.91 $\pm$ 4.21 \\ \hline  
Image & 2.82 $\pm$ 0.54  & 2.77 $\pm$ 0.51 & 2.74 $\pm$ 0.65 \\ \hline
Ringnorm & 4.41 $\pm$ 0.64 & 4.29 $\pm$ 0.69 & 5.05 $\pm$ 0.99 \\ \hline
Solar & 34.20 $\pm$ 1.75 & 34.27 $\pm$ 1.80 & 35.03 $\pm$ 1.89 \\ \hline 
Splice & 11.61 $\pm$ 0.81 & 11.85 $\pm$ 0.83 & 11.83 $\pm$ 0.80 \\ \hline 
Thyroid & 4.37 $\pm$ 2.19 & 4.20 $\pm$ 2.17 & 6.97 $\pm$ 3.78 \\ \hline
Titanic & 22.64 $\pm$ 1.34 & 22.50 $\pm$ 0.98 & 22.99 $\pm$ 2.81 \\ \hline
Twonorm & 3.05 $\pm$ 0.34 & 3.19 $\pm$ 0.51 & 3.43 $\pm$ 0.43 \\ \hline 
Waveform & 10.10 $\pm$ 0.48 & 9.95 $\pm$ 0.48 &  11.70 $\pm$ 0.88 \\ \hline
\end{tabular}
\end{sc}
\end{small}
\end{center}
\end{table}

\begin{table}
\caption{Data sets description: F-Measure Experiment. Here, $n$, $D$, $p$, $nr$ and $PPE$ represent the numbers of training examples, input dimension, test examples, train/test partitions and approximate percentage of positive examples respectively.}
\vskip 0.05in
\begin{small}
\begin{sc}
\begin{center}
\begin{tabular}{|l|c|c|c|c|c|} \hline
Dataset & $n$ & $D$  & $p$ & $nr$ & $PPE$\\ \hline
Yeast7  & 297 & 8 & 1187 & 50 & 2 \\ \hline 
Yeast5  & 320 & 8 & 1164 & 50 & 4 \\ \hline 
Car3  & 350 & 6 & 1378 & 50 & 4 \\ \hline 
Ecoli5 & 124 & 7 & 212 & 50 & 6 \\ \hline
Yeast4  & 165 & 8 & 1319 & 50 & 11 \\ \hline 
Breastcancer & 200 & 9 & 77 & 100 & 29 \\ \hline
German & 700 & 20 & 300 & 100 & 30 \\ \hline
Diabetes & 468 & 8 & 300 & 100 & 35 \\ \hline
\end{tabular}
\end{center}
\vskip -0.2in
\end{sc}
\end{small}
\end{table}

\begin{table}
\begin{center}
\caption{F-Measure Performance} 
\vskip 0.1in
\begin{small}
\begin{sc}
\begin{tabular}{|l|c|c|c|c|} \hline
 Dataset/Method & NLP & FM  &  NLP-FM(bias) \\ \hline
yeast7 & 32.24 $\pm$ 15.54 & 42.58 $\pm$ 7.84 &  40.85 $\pm$ 8.59 \\ \hline
yeast5 & 19.79 $\pm$ 12.85 & 32.85 $\pm$ 8.56 &  28.45 $\pm$ 10.57 \\ \hline
car3 & 55.89 $\pm$ 9.30 & 64.35 $\pm$ 8.51 & 62.67 $\pm$ 8.49 \\ \hline
ecoli5 & 84.41 $\pm$ 6.25 & 83.79 $\pm$ 5.90 &  84.08 $\pm$ 5.86 \\ \hline
yeast4 & 71.35 $\pm$ 3.95 & 73.64 $\pm$ 2.97 &  73.23 $\pm$ 2.81 \\ \hline
BreastCancer & 38.42 $\pm$ 10.55 & 47.34 $\pm$ 6.44  & 46.98 $\pm$ 6.37 \\ \hline
German & 54.15 $\pm$ 4.17 & 57.53 $\pm$ 2.95 &  56.91 $\pm$ 2.87 \\ \hline
Diabetes & 62.69 $\pm$ 3.49 & 66.23 $\pm$ 2.87 & 66.12 $\pm$ 2.67 \\ \hline
\end{tabular}
\end{sc}
\end{small}
\end{center}
\end{table}

\section{Experiments}
We conducted two experiments with various methods. See Table 2 for a summary of the various methods. In the first experiment we compared the performance of EP-CV(NLP) method with that of EP(ML) and LS-CV(NLP) methods. In the second experiment we compared the performance of EP-CV(FM) method with that of EP-CV(NLP) and two step classifier methods. We used the {\it minimize} Matlab routine of the GPML Matlab code available at {\it http://www.gaussianprocess.org/gpml/code/matlab/doc/} for hyperparameters optimization. In all the experiments we used a single global scaling hyperparameter.

\subsection{NLP Experiment}
In this experiment we used the thirteen benchmark datasets available in the web{\footnote {\it http://ida.first.fraunhofer.de/projects/bench/benchmarks.htm}} summarized in Table 3. Let us first consider the results from the first experiment given in Table 4 and Table 5. For the EP(ML) method we used the GPML Matlab code available in the web{\footnote {\it http://www.gaussianprocess.org/gpml/code/matlab/doc/}}.   

We conducted Friedman test \citep{Demsar:06} with the corresponding post-hoc tests for comparison of classifiers over multiple datasets. The comparison over multiple datasets requires a performance score of each method on each dataset \citep{Demsar:06}. Here, we consider the mean over the partitions of a given dataset as the performance score. As pointed out in \citet{Demsar:06}, it is not clear how to make use of the standard deviation information when the datasets are not independent over the partitions. The Friedman test ranks the methods for each dataset separately based on the chosen performance score (mean performance in our case). The best performing method gets the rank of 1, the second best rank 2 and so on. In case of ties, average ranks are assigned. The Friedman test checks whether the measured average ranks (over the datasets) are significantly different from the mean rank under the null hypothesis. Under the null hypothesis all the methods are equivalent and so their ranks should be equal. 

In the case of NLP performance measure, the measured average ranks for the EP(ML), EP-CV(NLP) and LS-CV(NLP) methods were 
$1.46$, $1.62$ and $2.92$ respectively. With three methods and 13 datasets, the F-statistic comparison at a significance level of 0.05 rejected the null hypothesis. Since the null hypothesis was rejected we conducted the Nemenyi post-hoc test for pairwise comparisons. This test revealed that the results of the EP(ML) and EP-CV(NLP) methods are better than the LS-CV(NLP) method at the significance level of $0.05$. On the other hand, the post-hoc test did not detect any significant difference in the results of EP(ML) and EP-CV(NLP) methods. In the case of test set performance measure, the measure averaged ranks for the EP(ML), EP-CV(NLP) and LS-CV(NLP) methods were $1.62$, $1.77$ and $2.62$ respectively. Note that the average rank of LS-CV method has improved on the test set error performance. Here again, the null hypothesis is rejected at the same significance level and the post-hoc test did not detect any significant difference in the results of EP(ML) and EP-CV(NLP) methods. The results of the EP(ML) and EP-CV(NLP) methods are better than the LS-CV(NLP) method at the significance level of $0.05$ and $0.1$ respectively. Thus, we can conclude that EP(ML) and EP-CV(NLP) are competitive to each other. Further, both these methods perform better than the LS-CV(NLP) method in this experiment.  

We can also make other observations from the tables. We note that the NLP performance of the LS-CV(NLP) method is quite inferior on several datasets even though its test set error performances on most of these datasets (except {\it waveform and thyroid}) are relatively closer. Further, some kind of group behavior can be seen. For example, the NLP scores are high on {\it titanic}, {\it breast-cancer}, {\it diabetes}, {\it German} and {\it flare-solar}. Also, the test set errors are $>20\%$ on these datasets. Consequently, we may consider these datasets as difficult ones. From Table 4 we observe that the NLP performance of LS-CV(NLP) is {\it closer} to the other two methods on these datasets (compared to its performance on other datasets). Next we can order the remaining datasets {\it heart}, {\it splice}, {\it banana}, {\it waveform}, {\it ringnorm}, {\it thyroid}, {\it twonorm} and {\it image} in terms of descending difficulty. Note that the difference in the NLP performance seems to have an increasing trend as the dataset becomes easier. To understand this we looked at the predictive probabilities of these methods on both correctly and wrongly classified examples. In the case of thyroid dataset these average probability scores for the LS-CV(NLP) method were $0.84$ (for correct classification) and $0.35$ (wrong classification). Here, the averaging was done over all the correctly(wrongly) classified examples over all the partitions. On the other hand, the corresponding values for the EP-CV(NLP) were $(0.96, 0.39)$. In the case of banana dataset, these scores were $(0.79, 0.35)$ and $(0.92, 0.29)$ for the LS-CV(NLP) and EP-CV(NLP) methods respectively. In the case of German dataset, they were $(0.75, 0.33)$ and $(0.79, 0.32)$. The scores for the EP-ML method were very close to that of the EP-CV(NLP) method. In general, we observed that the predictive probability estimates from the LS-CV(NLP) method were relatively poor and resulted in poor NLP performance. We looked at the hyperparameter estimates of the different methods and observed that the LS-CV(NLP) method takes smaller width and signal variance hyperparameter values on most of the datasets except on the difficult datasets (mentioned above) compared to the other two methods. Apart from this we did not observe any specific pattern in the hyperparameter values chosen by these methods. Looking at the hyperparameter estimates of EP(ML) and EP-CV(NLP), it seems that several solutions in the space of hyperparameters that give close performances are possible.       

\subsection{F-measure Experiment}
 The datasets used in this experiment are described in Table 6.  The datasets {\it yeast}, {\it car} and {\it ecoli} are multi-class datasets and we converted them into binary classification datasets by considering examples belonging to the class label indicated by the number (for example, $7$ in yeast7) as positive class respectively and treating the rest of the examples as negative class. These datasets are available in the web  {\footnote {\it ftp://ftp.ics.uci.edu/pub/machine-learning-databases/}}. We created 50 partitions for these datasets in a stratified manner reflecting the class distributions.

Let us consider the results from the second experiment given in Table 7. In this experiment all the results were obtained within the EP-CV framework. The first and second columns represent results obtained using NLP and smoothed F-measure (i.e., eq.~(\ref{FMmooth})) as the optimization criterion respectively. The third column represents results obtained from the two step classifier described earlier. We looked at the hyperparameter estimates of the different methods. We observed that the bias estimates of the two step classifier were somewhat closer (within $10\%$) to those of the smoothed F-measure method on the {\it breast-cancer}, {\it diabetes} and {\it German} datasets. On the remaining datasets they were different by more than $40\%$. The width and signal variance hyperparameter estimates were also quite different. From Table 7, we observe that the two step method is also good and gives closer performance to the smoothed F-measure method on several datasets. Further analysis of the performance results revealed that even though the standard deviations are high, the smoothed F-measure method gave better performance than the two step method on majority of the partitions on several datasets. We believe that the larger standard deviations in the results arises from the sensitivity to the dataset with lesser number of positive examples. To carry out the statistical significance tests, again we used the mean (over the partitions) as the performance score for each of the methods. The measured average ranks for these three methods were $2.75$, $1.25$ and $2.00$ respectively. With three methods and 8 datasets, the F-statistic comparison at a significance level of 0.05 rejected the null hypothesis. The Nemenyi post-hoc test for pairwise comparisons revealed that {\it only} the results of smoothed F-measure is better than the NLP based method at the significance level of 0.05. In this experiment the post-hoc test did not detect any significant differences in the comparisons of smoothed F-measure method with the two step method and the two step method with the NLP method. However in these two comparisons the rank differences were closer to the required critical differences at the significance level of 0.1.  We also observed that if we were to conduct Wilcoxon signed-rank test on these methods (as if we were comparing only two classifiers) then the results were statistically significant at the significance level of 0.05 for all the three pairs. In summary, the results demonstrate the usefulness of direct optimization of smoothed F-measure.
  
\section{Conclusion}
In this paper, we considered the problem of Gaussian process classifier model selection with different LOO-CV based optimization criteria and provided a practical algorithm using LOO predictive distributions with criteria like standard NLP, smoothed F-measure and WER to select hyperparameters. More specifically, apart from optimization of standard NLP, we demonstrated its usefulness in direct optimization of smoothed F-measure, which is useful to handle imbalanced data. We considered predictive distribution arrived from the Expectation Propagation (EP) approximation. We derived relevant expressions and proposed a very useful EP-CV algorithm. The experimental results on several real world benchmark datasets showed comparable NLP generalization performance (with NLP optimization) with existing approaches. We demonstrated that the smoothed F-measure optimization method is a very useful method that improves the F-measure performance significantly. Overall, the EP-CV algorithm is an excellent choice for GP classifier model selection with different LOO-CV based optimization criteria.         

\vskip 0.2in
\bibliography{cvgpc}
\section*{Appendix: Optimization criteria, approximate predictive distributions and their derivatives}
From the definitions of various optimization criteria like NLP measure (\ref{nlp}) and (\ref{FMmooth}) etc, we see that the chosen measure and its derivatives can be obtained from the LOO predictive distributions  $p(y_i|x_i,{\it S}_{\setminus{i}},\bftheta)$ and their derivatives ${{\partial p(y_i|x_i,{\it S}_{\setminus{i}},\bftheta)} \over {\partial \theta_j}}$. Here, $\theta_j$ is $j^{th}$ component of the hyperparameter vector $\bftheta$. Note that the derivative of the NLP measure (\ref{nlp}) is given by:
\begin{equation}
	{{\partial G(\bftheta)} \over {\partial \theta_j}}\:=\:- {1\over n} \sum_{i=1}^n {1\over \Phi(y_i z_i)} {{\partial p(y_i|x_i,{\it S}_{\setminus{i}},\bftheta)} \over {\partial \theta_j}}
\label{nlpderv}
\end{equation}
and the derivative of the smoothed F-measure (\ref{FMmooth}) is given by:
\begin{equation}
	 	{{\partial F_{\zeta}(\bftheta)} \over {\partial \theta_j}}\:=\:{{\eta(\bftheta) {{\partial A(\bftheta)} \over {\partial \theta_j}} - A(\bftheta) (1-\zeta) {{\partial m_+(\bftheta)} \over {\partial \theta_j}}} \over {\eta^2(\bftheta)}}
\label{FMoothderv}
\end{equation}
where $\eta(\bftheta)\:=\:\zeta n_{+} + (1-\zeta)m_{+}(\bftheta)$. Note that the derivatives ${{\partial A(\bftheta)} \over {\partial \theta_j}}$ and ${{\partial m_+(\bftheta)} \over {\partial \theta_j}}$ are directly dependent on ${{\partial p(y_i=+1|x_i,{\it S}_{\setminus{i}},\bftheta)} \over {\partial \theta_j}}$. Now, from (\ref{Pyi}) we see that to define the LOO predictive distributions, we need the LOO mean $\mu_{\setminus{i}}$ and variance $\sigma^2_{\setminus{i}}$. In the case of EP approximation, analytical expressions to compute these quantities are already available (see eqn. (\ref{cavmu})). Next, we give details on how the derivatives of predictive distributions can be obtained with these approximations.

\subsection*{Derivatives of Predictive Distributions}
For ease of reference we recall (\ref{Pyi}) here. $$p(y_i|x_i,{\it S}_{\setminus{i}},\bftheta)\:=\:\Phi \Bigl({{y_i(\mu_{\setminus{i}}+\gamma)} \over {\sqrt{1+\sigma^2_{\setminus{i}}}}}\Bigr).$$ Then, with $z_i\:=\:{{\mu_{\setminus{i}}+\gamma} \over {\sqrt{1+\sigma^2_{\setminus{i}}}}}$ and $N(z_i)\:=\:{1 \over \sqrt{2\pi}} \exp(-{{z^2_i} \over 2})$ we have $${{\partial p(y_i|{\bf x}_i,{\it S}_{\setminus{i}},\bftheta)} \over {\partial \theta_j}}\:=\:{{{\it N}(z_i) y_i} \over {\sqrt{1+\sigma^2_{\setminus{i}}}}}\Bigl({{\partial \mu_{\setminus{i}}}\over {\partial \theta_j}}\:-\:{1\over 2}{{z_i} \over  {\sqrt{1+\sigma^2_{\setminus{i}}}}} {{\partial \sigma^2_{\setminus{i}}} \over {\partial \theta_j}}\Bigr).$$ Here, $\theta_j$ represents any element of $\bftheta$ other than $\gamma$. Similarly, we have  
$${{\partial p(y_i|{\bf x}_i,{\it S}_{\setminus{i}},\bftheta)} \over {\partial \gamma}}\:=\:{{{\it N}(z_i) y_i} \over {\sqrt{1+\sigma^2_{\setminus{i}}}}}.$$  
Thus, we need ${\partial \mu_{\setminus{i}}}\over {\partial \theta_j}$ and ${\partial \sigma^2_{\setminus{i}}} \over {\partial \theta_j}$. Below, we give details on how they can be obtained with the EP approximation.    


\subsection*{Derivatives of LOO mean and variance with fixed site parameters: EP Approximation}
In the case of EP approximation, since we resort to EM type optimization, we derive expressions for ${{\partial \mu_{\setminus{i}}} \over {\partial \theta_j}}$ and ${{\partial \sigma^2_{\setminus{i}}} \over {\partial \theta_j}}$ assuming that the site parameters are fixed. From $\mu_{\setminus{i}}\:=\:\sigma^2_{\setminus{i}}\bigl({m_i \over {\bf C}_{ii}}-{\mu_i \over \sigma^2_i}\bigr)$, we have
$${{\partial \mu_{\setminus{i}}} \over {\partial \theta_j}}\:=\:{{\mu_{\setminus{i}}} \over {{\sigma^2}_{\setminus{i}}}} {{\partial \sigma^2_{\setminus{i}}} \over {\partial \theta_j}}\:+\:{{\sigma^2_{\setminus{i}}} \over {({\bf C}_{ii})^2}} \Bigl({\bf C}_{ii}{{\partial {\bf m}_i} \over {\partial \theta_j}}\:-\:{\bf m}_i {{\partial {\bf C}_{ii}} \over {\partial \theta_j}}\Bigr).$$ From $\sigma^2_{\setminus{i}}\:=\:(({\bf C}_{ii})^{-1}\:-\:\sigma^{-2}_i)^{-1}$, we have
$${{\partial \sigma^2_{\setminus{i}}} \over {\partial \theta_j}}\:=\: {{\sigma^4_{\setminus{i}}} \over {({\bf C}_{ii})^2}} {{\partial {\bf C}_{ii}} \over {\partial \theta_j}}.$$ Since ${\bf m}\:=\:{\bf C}{\bfSigma^{-1}}{\bfmu}$, we have ${{\partial {\bf m}} \over {\partial \theta_j}}\:=\:{{\partial {\bf C}} \over {\partial \theta_j}} {\bfSigma^{-1}}{\bfmu}$. Note that ${\bf C}$ can be re-written using Sherman-Morrison-Woodbury formula as: ${\bf C}\:=\:{\bf K}\:-\:{\bf K}({\bf K}\:+\:\bfSigma)^{-1}{\bf K}$ and it is useful to work with this expression to achieve improved numerical stability \citep{Raswil:06} as it avoids inversion of ${\bf K}$. Then we have 
$${{\partial {\bf C}} \over {\partial \theta_j}}=\:({\bf I}\:-\:({\bf K}+\bfSigma)^{-1}{\bf K})^T{{\partial {\bf K}} \over {\partial \theta_j}}({\bf I}\:-\:({\bf K}+\bfSigma)^{-1}{\bf K}).$$ Note that ${{\partial {\bf C}_{ii}} \over {\partial \theta_j}}$, $i \in {\tilde I}$ (where ${\tilde I}=\{1,2,\ldots,n\}$) are nothing but the diagonal entries of the above expression. Note also that ${{\partial {\bf m}} \over {\partial \theta_j}}$ can be efficiently computed by taking advantage of the presence of the vector ${\bfSigma^{-1}}{\bfmu}$. But, to compute ${{\partial {\bf C}_{ii}} \over {\partial \theta_j}}$, $i \in {\tilde I}$ we cannot avoid the matrix multiplication with ${{\partial {\bf K}} \over {\partial \theta_j}}$; this results in $O(n^3)$ for each $\theta_j$. Finally, it is useful to re-write $({\bf K}+\bfSigma)^{-1}\:=\:\bfSigma^{-{1\over2}}({\bf I}\:+\:\bfSigma^{-{1\over2}}{\bf K}\bfSigma^{-{1\over2}})^{-1}\bfSigma^{-{1\over2}}$ \citep{Raswil:06}.     

\end{document}